\documentclass[conference]{IEEEtran}
\IEEEoverridecommandlockouts

\usepackage{cite}
\usepackage{amsmath,amssymb,amsfonts}
\usepackage{algorithm}
\usepackage{algorithmic}
\usepackage{multirow}
\usepackage{booktabs}
\usepackage{graphicx}
\usepackage{textcomp}
\usepackage{xcolor}
\def\BibTeX{{\rm B\kern-.05em{\sc i\kern-.025em b}\kern-.08em
    T\kern-.1667em\lower.7ex\hbox{E}\kern-.125emX}}
\begin{document}

\title{Unifying VLM-Guided Flow Matching and Spectral Anomaly Detection for Interpretable \\ Veterinary Diagnosis}

\author{
    Pu Wang$^{1,2}$, Zhixuan Mao$^{1}$, Jialu Li$^{3}$, Zhuoran Zheng$^{4}$, Dianjie Lu$^{5}$, Youshan Zhang$^{6,*}$ \\
    \normalsize $^{1}$\textit{School of Mathematics, Shandong University},
    \normalsize $^{2}$\textit{Shenzhen Loop Area Institute},
    \normalsize $^{3}$\textit{Yeshiva University},\\
    \normalsize $^{4}$\textit{Qilu University of Technology},
    \normalsize $^{5}$\textit{Shandong Normal University},
    \normalsize $^{6}$\textit{Chuzhou University}, \\
    \normalsize wangou@mail.sdu.edu.cn, youshan\_zhang@chzu.edu.cn
  
}

\maketitle

\let\thefootnote\relax\footnotetext{* Corresponding author. This research was funded by the research project of Chuzhou University (Grant No. 2025qd36).}

\begin{abstract}
Automatic diagnosis of canine pneumothorax is challenged by data scarcity and the need for trustworthy models.   To address this, we first introduce a public, pixel-level annotated dataset to facilitate research.   We then propose a novel diagnostic paradigm that reframes the task as a synergistic process of signal localization and spectral detection.   For localization, our method employs a Vision-Language Model (VLM) to guide an iterative Flow Matching process, which progressively refines segmentation masks to achieve superior boundary accuracy.  For detection, the segmented mask is used to isolate features from the suspected lesion.   We then apply Random Matrix Theory (RMT), a departure from traditional classifiers, to analyze these features.   This approach models healthy tissue as predictable random noise and identifies pneumothorax by detecting statistically significant outlier eigenvalues that represent a non-random pathological signal.   The high-fidelity localization from Flow Matching is crucial for purifying the signal, thus maximizing the sensitivity of our RMT detector.   This synergy of generative segmentation and first-principles statistical analysis yields a highly accurate and interpretable diagnostic system
(source code is available at: https://github.com/Pu-Wang-alt/Canine-pneumothorax).
\end{abstract}


\section{Introduction}
Canine pneumothorax is a common and potentially life-threatening emergency in veterinary clinical practice characterized by abnormal accumulation of gas in the pleural space between the lungs and the chest wall, resulting in lung collapse and severe respiratory distress~\cite{jobson2016nursing}. Timely and accurate diagnosis is essential to guide emergency treatment and improve prognosis. At present, chest X-ray radiography is a common method for the diagnosis of canine pneumothorax. However, the interpretation of radiological images is highly dependent on the expertise and clinical experience of veterinarians. In some subtle or atypical cases, manual interpretation may be subjective, and in emergency situations, it is challenging to quickly and accurately delineate the extent of collapse for assessing the severity of the disease and making treatment plans (such as thoracocentesis). Therefore, it is of great clinical application value to develop an intelligent tool that can assist veterinarians in rapid, objective, and accurate diagnosis.
\begin{figure}[t!]
    \centering
    \includegraphics[width=1\linewidth]{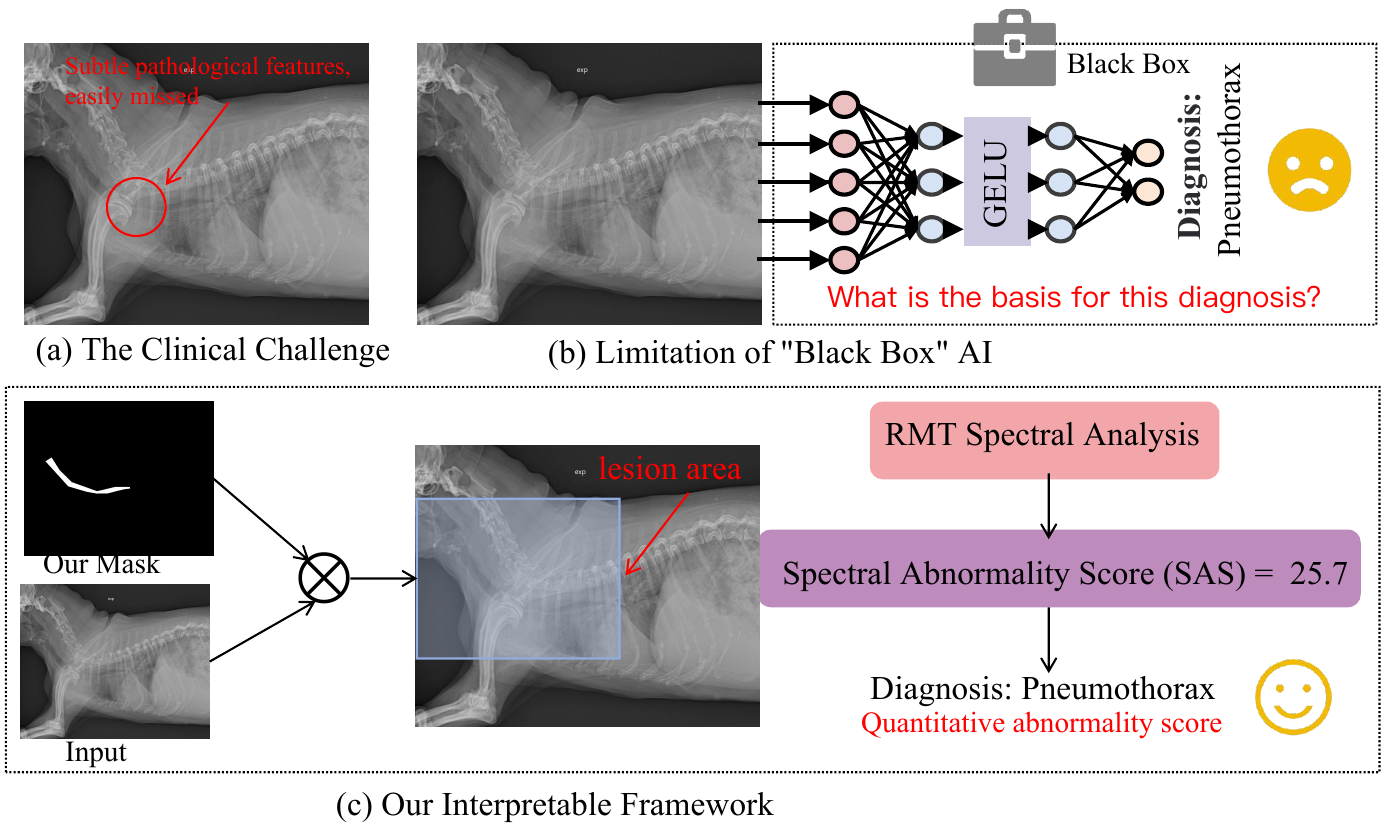}
    \caption{Comparison of diagnostic approaches for canine pneumothorax. (a) The clinical challenge of subtle features. (b) The interpretability issue of "black box" AI. (c) Our proposed framework.
    }
    \vspace{-3mm}
    \label{fig:1}
\end{figure}
Recently, artificial intelligence technology represented by deep learning has made breakthroughs in the field of medical image analysis, and shows great potential, especially in lesion segmentation and classification tasks~\cite{asgari2021deep,wang2026power}. In veterinary radiology, AI algorithms have been initially applied to tasks such as assessment of canine hip dysplasia~\cite{loureiro2025deep}, heart size measurement~\cite{ramisetty2024precision}, and identification of certain skeletal abnormalities~\cite{kostenko2024machine}, showing great potential for improving diagnostic objectivity and efficiency. However, these traditional AI methods face two major bottlenecks. One is the extreme scarcity of large-scale, high-quality labeled data. There is a serious lack of standardized public datasets with high-quality expert annotations in the field of veterinary imaging. The construction of such a dataset is not only costly, but also requires the time of a large number of veterinary radiology experts. The second is the lack of interpretability. 
As illustrated in Figure~\ref{fig:1}, traditional models often function as “black boxes" that usually only provide numerical results for segmentation or classification and are unable to explain their diagnostic rationale, which limits their application in clinical decision making where a high degree of trust is required. 
Furthermore, distinct from general human medical imaging, veterinary radiology faces the unique challenge of extreme interspecific and interbreed anatomical variance (e.g., the thoracic cavity shape differences between a Chihuahua and a Great Dane). This high variance makes it difficult for standard supervised methods to abstract a unified ``normal" representation, often leading to poor generalization. In contrast, our proposed framework addresses this by modeling the statistical properties of the signal rather than memorizing anatomical shapes, providing a transparent and trustworthy alternative by combining precise lesion localization with a quantitative anomaly score.

With the development of large-scale pre-trained Foundation Models, especially large language models (LLMS) and Vision-language models (VLMS)~\cite{zhang2024vision}, these models have gained unprecedented world knowledge and powerful zero-shot/few-shot inference capabilities through pre-training on massive multi-modal data~\cite{li2025vt,li2026dvla}. 
Their unique ability to understand and generate natural language opens up entirely new possibilities for building trustworthy human-computer interactive diagnostic systems~\cite{rane2023contribution}. Although LLM has shown great potential in the field of general human medicine, there is still a huge research gap in the highly specialized field of veterinary radiology.

To address the data scarcity problem, we begin by constructing and releasing the first publicly available radiological image dataset with pixel-level expert annotations for canine pneumothorax. Based on this foundation, we propose an innovative VLM-FlowMatch segmentation framework, semantically guided lesion localization by iteratively refining an initial segmentation mask with a VLM-guided vector field. Finally, for diagnostic task, we introduce a paradigm based on RMT for anomaly detection, which quantifies the statistical perturbation from pathological signals within the focused lesion area to provide a robust Spectral Anomaly Score (SAS).

\section{Related Work}
\textbf{Canine medical image segmentation.} Medical image segmentation is the cornerstone of computer-aided diagnosis, which aims to accurately identify anatomical structures and lesion regions at the pixel level~\cite{cui2023advances}. Fully supervised deep learning models, represented by U-Net and its variants, have achieved outstanding achievements in numerous segmentation tasks and become the gold standard in this field~\cite{ronneberger2015u}. However, the success of these models is premised on large-scale, high-quality pixel-level labeled data. In specialized fields such as veterinary radiology, the cost of obtaining such data is extremely high, severely limiting the application of fully supervised methods~\cite{xiao2025review}. To address this challenge, the research community has explored a variety of data-efficient learning strategies, aiming to learn more robust features from limited labeled data. These methods include transfer learning~\cite{kim2022transfer}, weakly supervised learning~\cite{ren2023weakly}, and advanced techniques based on feature matching and distribution alignment~\cite{huang2024survey}.
Although these data-efficient methods effectively alleviate the problem of data dependence, the trained models still face two major limitations: (1) an accuracy bottleneck in fuzzy and subtle boundaries; (2) a lack of explanations for the diagnosis.

\textbf{Applications of Large Language Models in Medical Imaging.} In recent years, large language models (LLMS) and vision-language models (VLM) have brought advances to the field of medical image analysis~\cite{wang2025agentpolyp}. Although traditional deep learning models perform well on tasks such as classification or segmentation, their nature of not being able to communicate effectively with clinicians has been a major obstacle in their clinical translation. LLM has advanced logical reasoning and natural language interaction capabilities, which can transform complex pixel information into language that human doctors can understand and verify~\cite{li2024snapkv}. In Visual Question answering (VQA) and diagnostic AIDS, models are able to respond to natural language questions (such as Are there abnormalities in the image? ) to answer the specific content of the image, and even directly give preliminary diagnosis and classification recommendations~\cite{bazi2023vision}. In the automatic generation of radiology reports, the model automatically analyzes the input medical images and generates a structured and standardized diagnostic report, which can reduce the work burden of radiologists and standardize the quality of the report~\cite{alfarghaly2021automated}. However, the reliability and factual accuracy of the model are still huge challenges, and sometimes it will produce plausible but inconsistent illusion~\cite{scire2024truth}. Most studies use LLM as an isolated, end-of-process module that lacks intervention and insight into upstream image processing steps such as segmentation.

\begin{figure*}[ht!]
    \centering
    \includegraphics[width=1\linewidth]{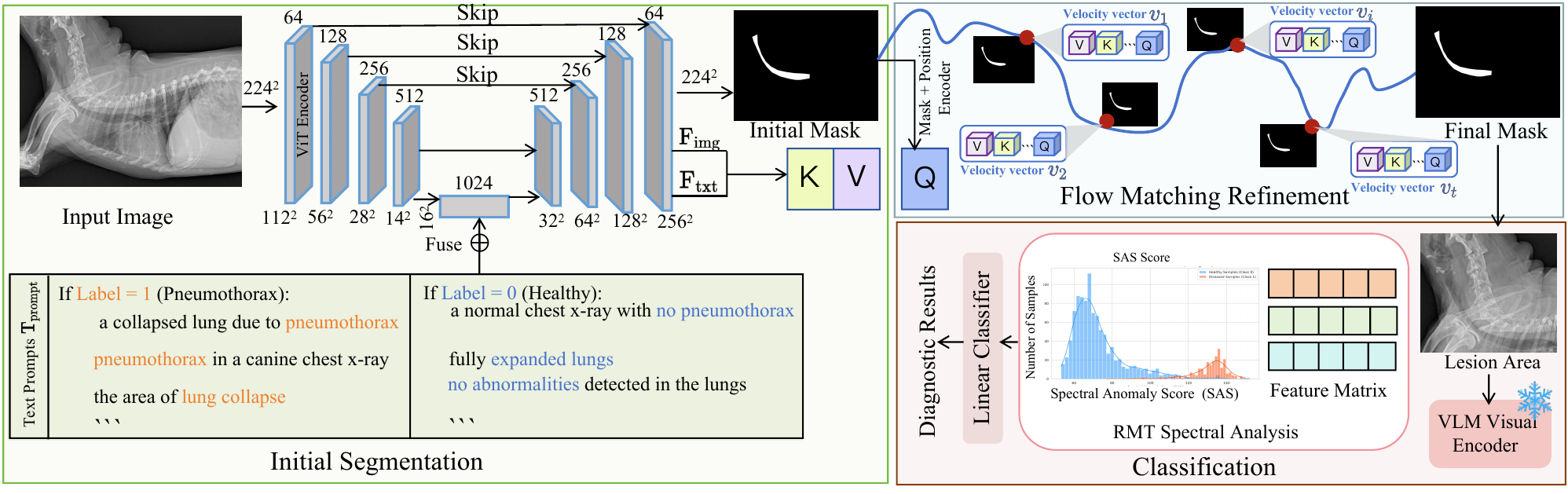}
    \caption{Overview of our proposed synergistic framework for canine pneumothorax diagnosis.}
    \label{fig:2}
\end{figure*}

\section{Method}
VLM-FlowMatch reframes canine pneumothorax diagnosis as a unified signal localization and spectral analysis process.
In Figure \ref{fig:2}, the pipeline first employs a VLM-Infused U-Net and an Attentional Flow Matching module to generate a high-precision segmentation mask $\hat{\text{M}}$. This mask acts as a spatial filter to isolate the region of interest (ROI). Finally, features from the focused region are analyzed via a Random Matrix Theory (RMT)-based classifier to render a final diagnosis.

\subsection{ViT-UNet for Initial Mask Generation}
Our architecture features a U-Net with a pre-trained Vision Transformer (ViT) backbone to leverage its global feature extraction for semantic understanding.   For an input image X, the ViT encoder generates a visual feature map $\text{F}_{\text{img}}$.   To incorporate semantic guidance, we perform channel-wise multiplication between $\text{F}_{\text{img}}$ and a projected text feature vector $\text{F}_{\text{txt}}$ (derived from prompt $\text{T}_{\text{prompt}}$), using spatial broadcasting for dimension alignment.

We construct the U-Net's skip connections by progressively upsampling this final text-fused feature map via bilinear interpolation. This forms a feature pyramid that matches the resolution of each decoder stage. The decoder then reconstructs the initial segmentation mask $\text{M}^{(0)}$:
\begin{equation}
    \text{M}^{(0)}, \text{F}_{\text{img}}, \text{F}_{\text{txt}} = \Psi(\text{X}, \text{T}_{\text{prompt}}; \theta_{\text{vlm-unet}})
\end{equation}

\subsection{Iterative Refinement via VLM-Guided Flow Matching}

To refine $\text{M}^{(0)}$, we learn a vector field $v$ through flow matching. We model the refinement as a discretization of an ordinary differential equation (ODE): $dx_t = v(x_t, t, \text{F}_{\text{cond}}) dt$.
At each timestep $t$, the network predicts the velocity $v_t$ using a cross-attention mechanism. Cross-attention injects multimodal context to predict boundary-correcting velocities:
\begin{equation}
\begin{split}
    v_t = \Phi_{\text{flow}}(\text{CrossAttention}(&Q=f(x_t), \\
    K=[\text{F}_{\text{txt}}; \text{F}_{\text{img}}], V=[\text{F}_{\text{txt}}; \text{F}_{\text{img}}]))
\end{split}
\end{equation}
Here, the conditioning term $\text{F}_{\text{cond}}$ is implicitly handled by the $K, V$ projection.

Training and Inference:
We construct a probability path $x_t = (1-t)x_0 + t x_1$ between the coarse mask $x_0 = \text{M}^{(0)}$ and the ground truth $x_1 = \text{M}_{gt}$. The network $v_\theta$ is trained with Conditional Flow Matching (CFM) loss to regress the target velocity $u_t = x_1 - x_0$.
During inference, we solve the ODE using an Euler solver with $T=10$ steps ($dt=0.1$). Starting from $x_0 = \text{M}^{(0)}$, we iterate to obtain $x_1$, which is then binarized to yield the final mask $\hat{\text{M}}$.The comprehensive workflow of this guidance process is outlined in Algorithm~\ref{alg:refinement}.

\subsection{Spectral Anomaly Detection for Diagnostic Classification}

We utilize Random Matrix Theory (RMT) to detect pathological signals. We isolate the ROI by $\text{X}_{\text{focus}} = \text{X} \odot \hat{\text{M}}$. The features extracted from $\text{X}_{\text{focus}}$, denoted as $\text{F}_p \in \mathbb{R}^{N \times p}$, are standardized to zero mean and unit variance.

Hypothesis Testing with RMT:
Under the null hypothesis ($H_0$), we assume the standardized features of healthy tissue approximate a high-dimensional random noise system. According to the Marchenko-Pastur (MP) law, as $N, p \to \infty$ with aspect ratio $p/N \to y$, the eigenvalues of the sample covariance matrix $S = \frac{1}{N}\text{F}_p^T \text{F}_p$ should fall within the support $[\lambda_{-}, \lambda_{+}]$, where $\lambda_{\pm} = (1 \pm \sqrt{y})^2$.

Under the alternative hypothesis ($H_1$), pneumothorax introduces a low-rank structural signal $U$, modeling the covariance as a "spiked" model: $\text{F}_p = W + U$. This causes outlier eigenvalues to separate from the MP bulk spectrum ($\lambda_i > \lambda_{+}$). We quantify this using the \textbf{Spectral Anomaly Score (SAS)}:
\begin{equation}
    \text{SAS}(\text{X}_{\text{focus}}) = \sum_{\lambda_i > \lambda_{+}} (\lambda_i - \lambda_{+})
\end{equation}

\begin{algorithm}[t]
\caption{VLM Guided Flow Matching Refinement}
\label{alg:refinement}
\begin{algorithmic}[1]
\REQUIRE Image $X$, Text Prompt $T_{\text{prompt}}$, Initial Mask $M^{(0)}$, Steps $N_{\text{steps}}$
\ENSURE Refined Mask $\hat{M}$
\STATE \textbf{Feature Extraction:}
\STATE $F_{\text{img}}, F_{\text{txt}} \leftarrow \text{VLM\_Encoder}(X, T_{\text{prompt}})$
\STATE \textbf{Initialization:}
\STATE $x_0 \leftarrow M^{(0)}$ \COMMENT{Start from coarse mask}
\STATE $dt \leftarrow 1 / N_{\text{steps}}$
\FOR{$k = 0$ \textbf{to} $N_{\text{steps}} - 1$}
    \STATE $t \leftarrow k \times dt$
    \STATE \textbf{Construct Query:}
    \STATE $Q \leftarrow \text{FeatureExtract}(x_t)$
    \STATE \textbf{VLM Guidance (Cross Attention):}
    \STATE $v_t \leftarrow \Phi_{\text{flow}}(\text{CrossAttn}(Q, K=[F_{\text{txt}}; F_{\text{img}}], V=[F_{\text{txt}}; F_{\text{img}}]))$
    \STATE \textbf{ODE Solver Step (Euler):}
    \STATE $x_{t+dt} \leftarrow x_t + v_t \times dt$
    \STATE $x_{t+1} \leftarrow x_{t+dt}$
\ENDFOR
\STATE $\hat{M} \leftarrow \text{Binarize}(x_1)$
\RETURN $\hat{M}$

\end{algorithmic}
\end{algorithm}
\vspace{-1mm}
\subsection{Optimization Objective}

We employ a staged training strategy. The segmentation network is optimized using a hybrid loss $\mathcal{L}_{\text{seg}} = \mathcal{L}_{\text{Dice}} + \lambda_{\text{bce}}\mathcal{L}_{\text{BCE}}$.
For diagnosis, the scalar SAS is fed into a logistic regression classifier $\Psi_{\text{clf}}$. To address class imbalance, $\Psi_{\text{clf}}$ is trained using \textbf{Focal Loss}:
\begin{equation}
    \mathcal{L}_{\text{Focal}}(p_t) = -\alpha_t (1-p_t)^\gamma \log(p_t)
\end{equation}

\section{Experiments}

\subsection{Dataset}
The dataset used in our study was sourced from the public Canine Thoracic Radiograph collection available on the Korean AI-Hub platform (https://aihub.or.kr/). To guarantee the reproducibility of our research, we partitioned this curated dataset into fixed training, validation, and test sets.     Specifically, the training set comprises 8641 images, the validation set comprises 2468 images, and the test set comprises 1236 images.     All model training and evaluation reported in this paper were conducted on this fixed partition to ensure fair and comparable results.

The dataset exhibits a characteristic class imbalance consistent with medical screening scenarios. Specifically, the distribution follows an approximate \textbf{8:2 (4:1)} ratio, where healthy cases constitute roughly 80\% of the data and pneumothorax cases account for the remaining 20\%. This uneven distribution necessitates the use of the \textbf{F1-score} as the primary metric for robust evaluation and motivates our adoption of the \textbf{Focal Loss} to prevent the model from biasing towards the majority class.

To ensure the high quality and clinical relevance of the annotations, we established a rigorous consensus protocol.  The raw data were annotated by three board-certified veterinary radiologists.  In cases of disagreement regarding lesion boundaries, a senior radiologist reviewed and adjudicated the final ground truth.  


\subsection{Implementation Details}
Our framework was implemented using PyTorch. The UnetFlowMatch model was trained on our training set using the Adam optimizer with an initial learning rate of 1e-4. We utilized OpenClip as the evaluation and feedback model. The prompts for the VLM were carefully designed to elicit structured refinement instructions.  
We utilized OpenCLIP (ViT-B/32).   The visual encoder was frozen to preserve pre-trained knowledge, while the text encoder generated embeddings for the prompt: “A canine chest X-ray showing [pulmonary markings/pneumothorax]".
The VLM feature dimension is $p=512$. The sample size $N$ corresponds to the number of patches in the focused region (typically $N \approx 196$ for $224\times224$ input).
The same VLM was used for diagnostic classification. 
All experiments were conducted using two NVIDIA RTX 4090 GPUs (24GB each).

\subsection{Quantitative results}
\subsubsection{Comparison on Segmentation Performance}
Table~\ref{tab:seg_results} presents a comprehensive performance comparison between our method and a variety of state-of-the-art segmentation models. 
Our model consistently ranks first across all metrics on both the validation and test sets. 
Note that models like SAM and Swin-Transformer perform poorly (low mIoU). This is likely due to the significant domain shift between their pre-training data (natural images) and veterinary X-rays, causing them to segment the entire lung field rather than the specific pathological air pocket.
Specifically, on the test set, our method achieves a top mDice of 0.8953 and mIoU of 0.8114. This performance not only surpasses classic U-Net-based architectures like PolypFlow (0.8019 mIoU) and powerful Transformer-based models like DeepLabv3+ (0.7733 mIoU), but also significantly outperforms other recent Mamba-based approaches such as Swin-UMamba (0.7820 mIoU). The consistent lead on both validation and test sets also suggests a strong generalization ability of our model. These results robustly validate the superiority of our proposed framework with its VLM-guided module.

\begin{table}[h!]
\centering
\caption{Performance comparison with state-of-the-art segmentation methods.}
\label{tab:seg_results}
\begin{tabular}{cllcccc}
\hline
\multirow{2}{*}{Category} & \multirow{2}{*}{Year} & \multirow{2}{*}{Model} & \multicolumn{2}{c}{Test Set} \\
\cmidrule(lr){4-5}  
& & & mDice$\uparrow$ & mIoU$\uparrow$ \\ \hline

\multirow{5}{*}{Unet-based} & 2015& Unet~\cite{ronneberger2015u} & 0.8774 & 0.7878 \\
& 2018 & Unet++~\cite{zhou2018unet++} & 0.8712 & 0.7788 \\
 & 2025& PolypFlow~\cite{wang2025polypflow} & 0.8869 & 0.8019 \\
 & 2020& $U^2$Net~\cite{qin2020u2} & 0.8834 & 0.7965 \\
 & 2022& Swin-UNet~\cite{cao2022swin} & 0.8462 & 0.7424 
 \\
 & 2025& H-vmunet~\cite{wu2025h} & 0.8281 & 0.7147 
\\ \hline
\multirow{4}{*}{Others} & 2020& HRNet~\cite{wang2020deep} & 0.8780 & 0.7883 \\
 & 2017& SegNet~\cite{badrinarayanan2017segnet} & 0.8777 & 0.7880 \\
 & 2020& ResUnet~\cite{diakogiannis2020resunet} & 0.8670 & 0.7727 \\
& 2023 & SAM~\cite{kirillov2023segment} & 0.6731 & 0.5277 \\
 \hline
\multirow{2}{*}{Transformer-based} & 2018& DeepLabv3+~\cite{chen2018encoder}& 0.8681 & 0.7733 \\
 & 2021& Swin-Transformer~\cite{liu2021swin} & 0.5763 & 0.4165 \\ \hline
\multirow{2}{*}{Mamba-based} & 2024& Mamba-UNet~\cite{wang2024mamba} & 0.8506 & 0.7481 \\
 & 2024& Swin-UMamba~\cite{liu2024swin} & 0.8733 & 0.7820 \\ \hline
 
  \multicolumn{3}{c}{Ours} & \textbf{0.8953} & \textbf{0.8114} \\ \hline
\end{tabular}
\end{table}

\subsubsection{Comparison on Diagnostic Classification Performance}
We evaluated our framework on the diagnostic classification task against a comprehensive suite of baseline models \cite{szegedy2015going,simonyan2014very,he2016deep,huang2017densely,szegedy2016rethinking,chollet2017xception,szegedy2017inception,zoph2018learning,tan2019efficientnet,dosovitskiy2020image,wu2021cvt,bao2021beit}, as summarized in Table~\ref{tab:cls_results}. 
To ensure a strictly fair comparison, all baseline models were trained and evaluated on the same focused (masked) input data as our method. Specifically, the regions of interest were isolated using the segmentation masks, and the background was zeroed out for all classifiers. A key challenge of our dataset is class imbalance, making the F1-score the primary metric for robust evaluation. The impact of this imbalance is evident in baselines such as VGG16, which exhibits 100\% Recall but extremely low Precision (0.2451), indicating convergence to a trivial solution of over-predicting the positive class. In contrast, the results clearly highlight the superiority of our proposed method: it avoids this failure mode, achieving the highest accuracy of 0.9032 and, more importantly, a balanced top-ranking F1-score of 0.7962.

\begin{table}[ht!]
\centering
\caption{Performance comparison on the validation and test sets.}
\label{tab:cls_results}
\begin{tabular}{lcccc}
\hline
\multirow{2}{*}{Model} & \multicolumn{4}{c}{Test Set} \\ 
\cmidrule(lr){2-5}

& Acc.$\uparrow$ & Prec.$\uparrow$ & Rec.$\uparrow$ & F1$\uparrow$ \\ \hline


GoogleNet~\cite{szegedy2015going} &0.8270 & 0.6787 & 0.5587 & 0.6129 \\
VGG16~\cite{simonyan2014very} & 0.2451 & 0.2451 & \textbf{1.0000} & 0.3938 \\
ResNet50~\cite{he2016deep} & 0.5908 & 0.3494 & 0.7769 & 0.4821 \\
DenseNet201~\cite{huang2017densely} & 0.8379 & 0.7263 & 0.5438 & 0.6219 \\
Inceptionv3~\cite{szegedy2016rethinking} & 0.8485 & 0.7680 & 0.5471 & 0.6390 \\
Xception~\cite{chollet2017xception} & 0.8582 & 0.8312 & 0.5289 & 0.6465 \\
InceptionResnetV2~\cite{szegedy2017inception} & 0.8687 & \textbf{0.8914} & 0.5289 & 0.6639 \\
NasnetLarge~\cite{zoph2018learning} & 0.8424 & 0.7109 & 0.6017 & 0.6517 \\
EfficientNetB7~\cite{tan2019efficientnet} & 0.8233 & 0.6294 & 0.6793 & 0.6534 \\
Vision Transformer~\cite{dosovitskiy2020image} & 0.6370 & 0.3441 & 0.5306 & 0.4174 \\
CONVT~\cite{wu2021cvt} & 0.4457 & 0.2858 & 0.8413 & 0.4267 \\
Beit\_large~\cite{bao2021beit} & 0.6191 & 0.3603 & 0.7140 & 0.4789 \\ \hline
Ours & \textbf{0.9032} & 0.8222 & 0.7719 & \textbf{0.7962} \\
\hline
\end{tabular}
\end{table}


VGG16 and CONVT show high recall but suffer from very low precision, indicating a tendency to over-predict the positive class. Conversely, models like InceptionResNetV2 achieve high precision (0.8914) but at the expense of lower recall (0.5289). Our method, however, attains a strong balance, achieving a high precision of 0.8222 while maintaining a competitive recall of 0.7719. 


\begin{figure}
    \centering
    \includegraphics[width=1\linewidth]{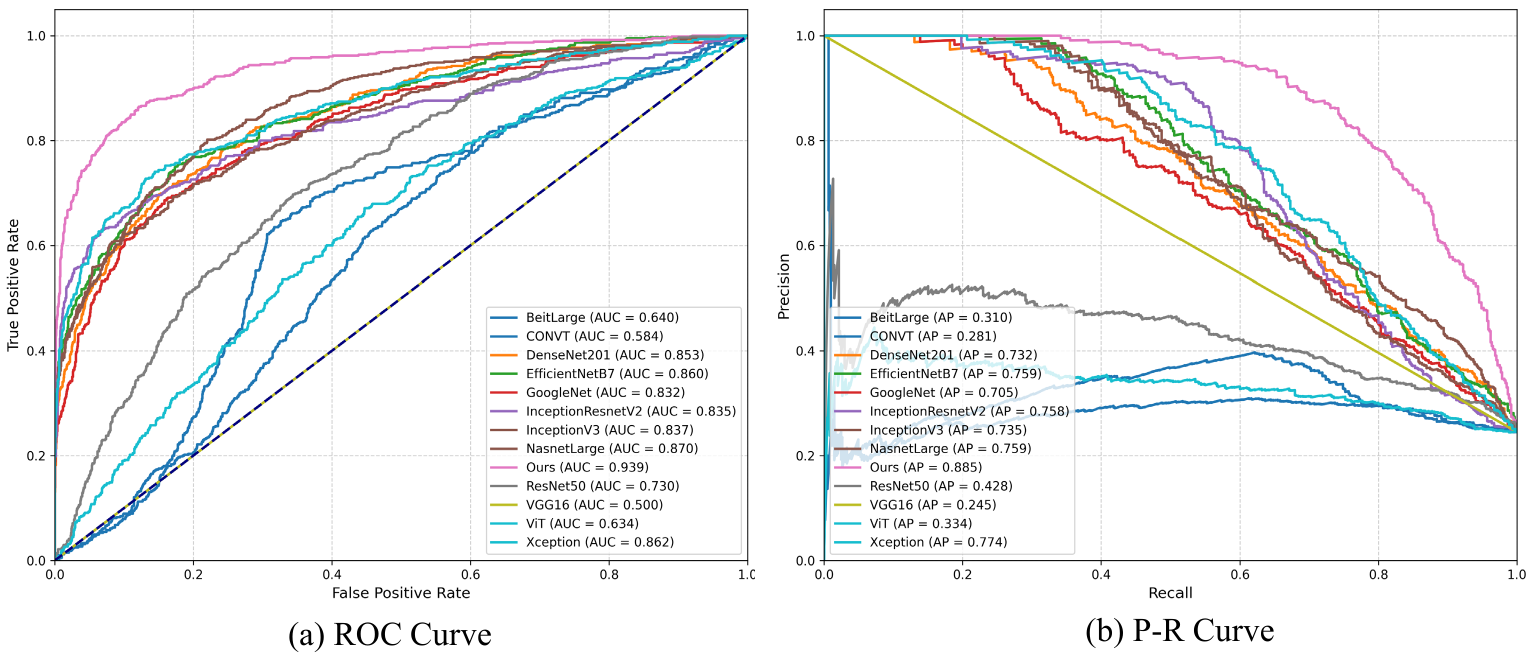}
    \caption{(a) This chart displays the Receiver Operating Characteristic curves. The proposed model achieves the best performance with a leading Area Under the Curve (AUC) score of 0.939. This is notably higher than other models. (b) This chart displays the Precision-Recall curves. The proposed model again shows superior performance, attaining the highest Average Precision score of 0.885.}
    \label{fig:placeholder2}
    \vspace{-3mm}
\end{figure}

\begin{table*}[t!]
\centering
\caption{Ablation study of our proposed framework. }
\label{tab:ablation_study}
\setlength{\tabcolsep}{+4.5mm}{
\begin{tabular}{@{}cccccccccc@{}}
\toprule
\multicolumn{1}{c}{Setting} & \multicolumn{3}{c}{Model Components} & \multicolumn{2}{c}{Seg. Performance} & \multicolumn{2}{c}{Class. Performance} \\
\cmidrule(r){2-4} \cmidrule(r){5-6} \cmidrule(l){7-8}
 & \begin{tabular}[c]{@{}c@{}}Text \\ Guidance\end{tabular} & \begin{tabular}[c]{@{}c@{}}Flow \\ Matching\end{tabular} & \begin{tabular}[c]{@{}c@{}}RMT Input \\ (Purification)\end{tabular} & mDice $\uparrow$ & mIoU $\uparrow$ & AUC $\uparrow$ & F1-Score $\uparrow$ \\ \midrule
\multicolumn{8}{l}{Experiment 1: Ablation on Segmentation Components} \\
(a) & $\times$ & $\times$ & -- & 0.8830  & 0.7949 & -- & -- \\
(b) & \checkmark & $\times$ & -- & 0.8736 & 0.7051 & -- & -- \\
(c) & \checkmark & \checkmark & -- & \textbf{0.8953 } & \textbf{0.8114} & -- & -- \\ \midrule
\multicolumn{8}{l}{Experiment 2: Ablation on Classification Synergy} \\
(d) & \checkmark & \checkmark & Full Image & -- & -- & 0.9054 & 0.7209 \\
(e) & \checkmark & \checkmark & Focused Image & -- & -- & \textbf{0.9390} & \textbf{0.7962} \\ \bottomrule
\end{tabular}}%
\end{table*}

To further assess the model's performance across all classification thresholds, we plotted the ROC and Precision-Recall (P-R) curves, as shown in Figure~\ref{fig:placeholder2}. In the ROC analysis (Figure~\ref{fig:placeholder2}a), our model achieves a superior AUC of 0.939, indicating its strong overall discriminative ability. More importantly, given the class imbalance of our dataset, the P-R curve (Figure~\ref{fig:placeholder2}b) provides a more insightful evaluation. Our model again leads with the highest Average Precision of 0.885. Its P-R curve is positioned consistently above all others, demonstrating a robust ability to maintain high precision even as recall increases. Both metrics confirm the comprehensive superiority of our proposed framework. High-resolution images are shown in the GitHub link.

\begin{figure}[h!]
    \centering
    \includegraphics[width=1\linewidth]{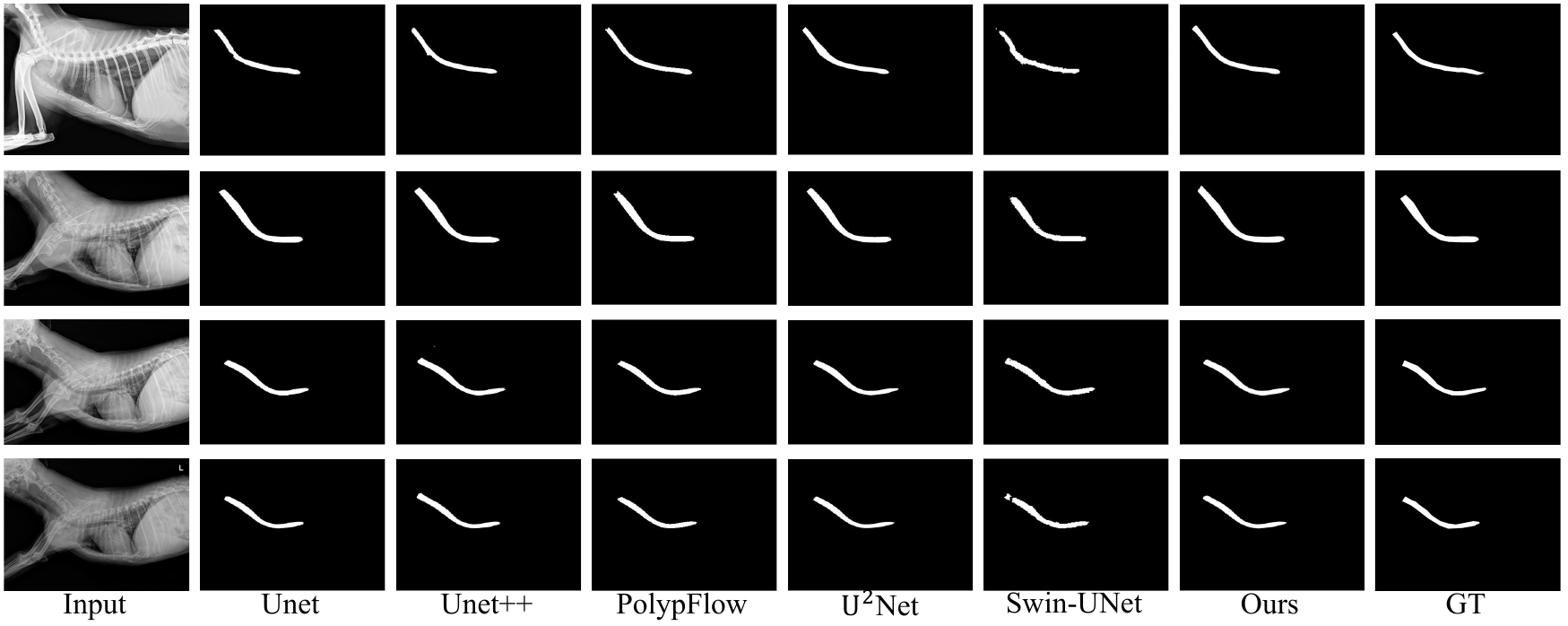}
    \caption{\textbf{Qualitative Comparison of Unet-based Segmentation Results.} This figure presents a visual comparison of the segmentation performance of our proposed model against five Unet-based methods.}
    \label{fig:seg 1}
    
\end{figure}

\subsubsection{Validation of RMT Assumptions: Empirical Spectral Distribution Analysis}
To empirically validate the core premise of our method—specifically that healthy tissue features follow the Marchenko-Pastur (MP) law while pneumothorax introduces outlier ``spikes". We visualized the Empirical Spectral Distribution (ESD) of the feature covariance matrices extracted from our test set.

\subsection{Qualitative Results}
To visually substantiate our quantitative findings, we provide qualitative comparisons of the segmentation results. As shown in Figure~\ref{fig:seg 1}, while U-Net-based models can capture the general shape of the target, our method produces cleaner boundaries and more accurate contours. 
In contrast, our model robustly and accurately segments the target structure in all cases. These visualizations are in strong agreement with our superior quantitative metrics and demonstrate the practical effectiveness of our approach.


\subsection{Ablation Study}
We conducted a series of ablation studies to validate the effectiveness of our framework's key components, with the results presented in Table~\ref{tab:ablation_study}.

Our segmentation ablation reveals that adding only VLM text guidance (b) degrades the baseline (a) performance.  However, the Flow Matching module (c) is crucial for refining this raw guidance, creating a synergistic effect that significantly surpasses the baseline with an mIoU of 0.8114.
The value of segmentation for classification is clear: focusing the input on the segmented lesion improved the F1-score from 0.7209 (full image) to 0.7962, confirming a strong synergistic benefit.

\section{Conclusion}
In this work, we introduce a novel, interpretable framework for canine pneumothorax diagnosis and release the first accompanying public, pixel-level annotated dataset. Our method uniquely unifies VLM-guided Flow Matching for precise lesion localization with Random Matrix Theory (RMT) for diagnosis, reframing the task as the detection of statistical anomalies in purified pathological signals. This synergistic paradigm is proven to significantly outperform state-of-the-art models, offering a new path for developing trustworthy medical AI in data-scarce environments.

\bibliographystyle{IEEEbib}
\bibliography{icme2026references}

\clearpage
\section*{Supplementary Material}

\setcounter{figure}{0}
\renewcommand{\thefigure}{S\arabic{figure}}

\setcounter{table}{0}
\renewcommand{\thetable}{S\arabic{table}}

\setcounter{algorithm}{0}
\renewcommand{\thealgorithm}{S\arabic{algorithm}}

\begin{figure}[htbp]
    \centering
    \includegraphics[width=1\linewidth]{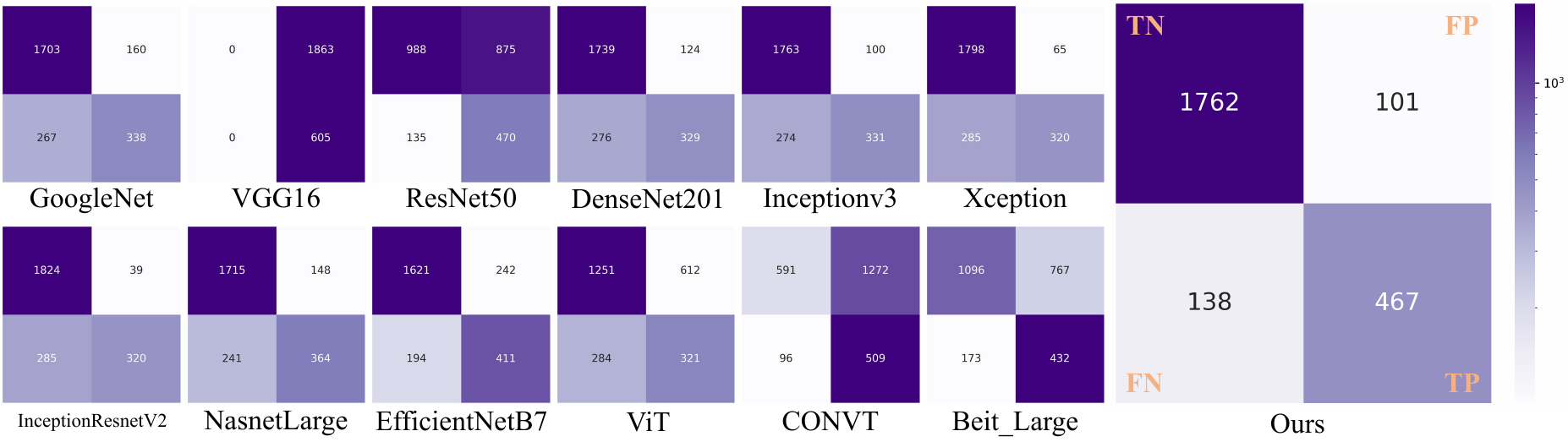}
    \caption{This figure presents a comparative analysis of the confusion matrices for the proposed model and twelve other models. Each matrix displays the counts for True Negatives (TN), False Positives (FP), False Negatives (FN), and True Positives (TP). The results highlight the superior performance of our model, which achieves a strong balance in correctly identifying both positive and negative instances while maintaining low error rates compared to the other methods.}
    \label{fig:s_confusion}
\end{figure}

To offer a more granular analysis, Figure~\ref{fig:s_confusion} displays the confusion matrices for all compared methods. The heatmap for our model (bottom right) provides a clear visualization of its balanced performance. It correctly identified 1762 negative cases (TN) and 467 positive cases (TP). More importantly, the number of misdiagnoses (False Positives, FP=101) and missed diagnoses (False Negatives, FN=138) are both effectively suppressed. This contrasts sharply with models like InceptionResnetV2, which, despite having very few FPs (FP=59, indicating a low rate of misdiagnosing healthy cases), missed a significant number of positive cases (FN=283), posing a high risk of missed diagnosis. Our framework's ability to minimize both FN and FP demonstrates its robustness and clinical potential in handling imbalanced diagnostic data, achieving an optimal balance between identifying patients and avoiding false alarms.

\begin{figure}[htbp]
    \centering
    \includegraphics[width=0.8\linewidth]{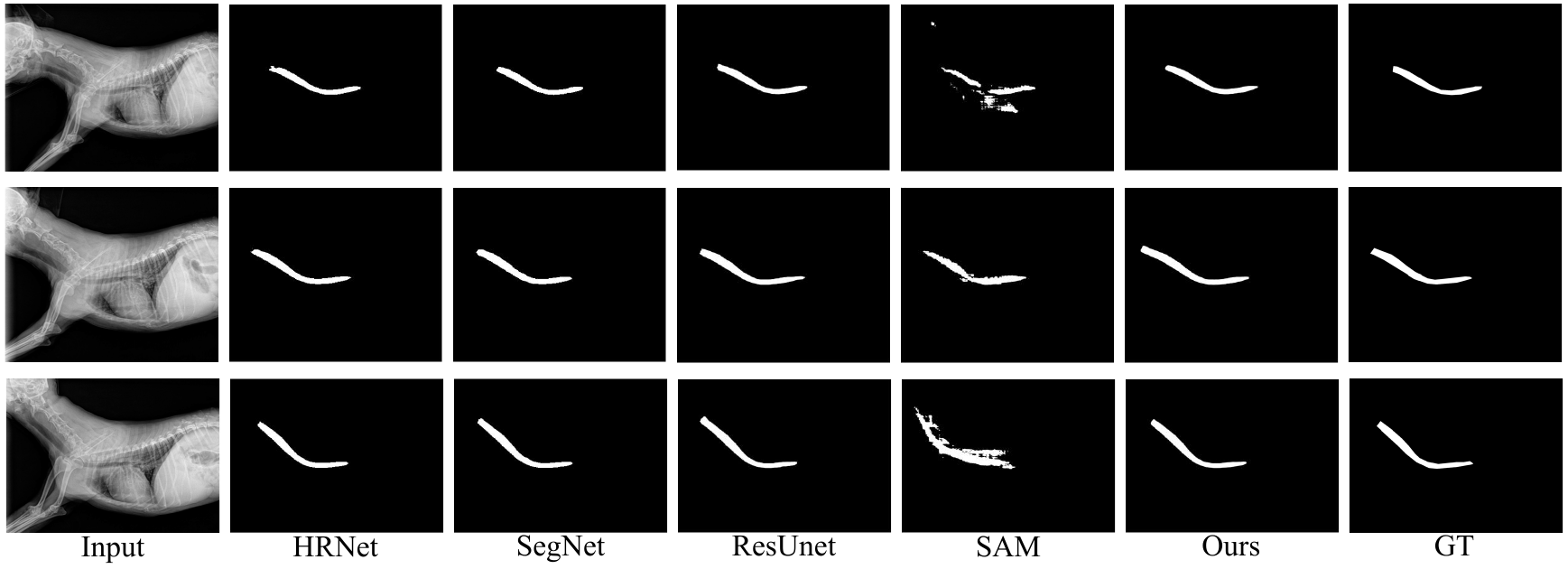}
    \caption{\textbf{Qualitative Comparison of Segmentation Results.} This figure presents a visual comparison of the segmentation performance of our proposed model against four methods.}
    \label{fig:s_seg2}
\end{figure}

\begin{figure}[htbp]
    \centering
    \includegraphics[width=0.8\linewidth]{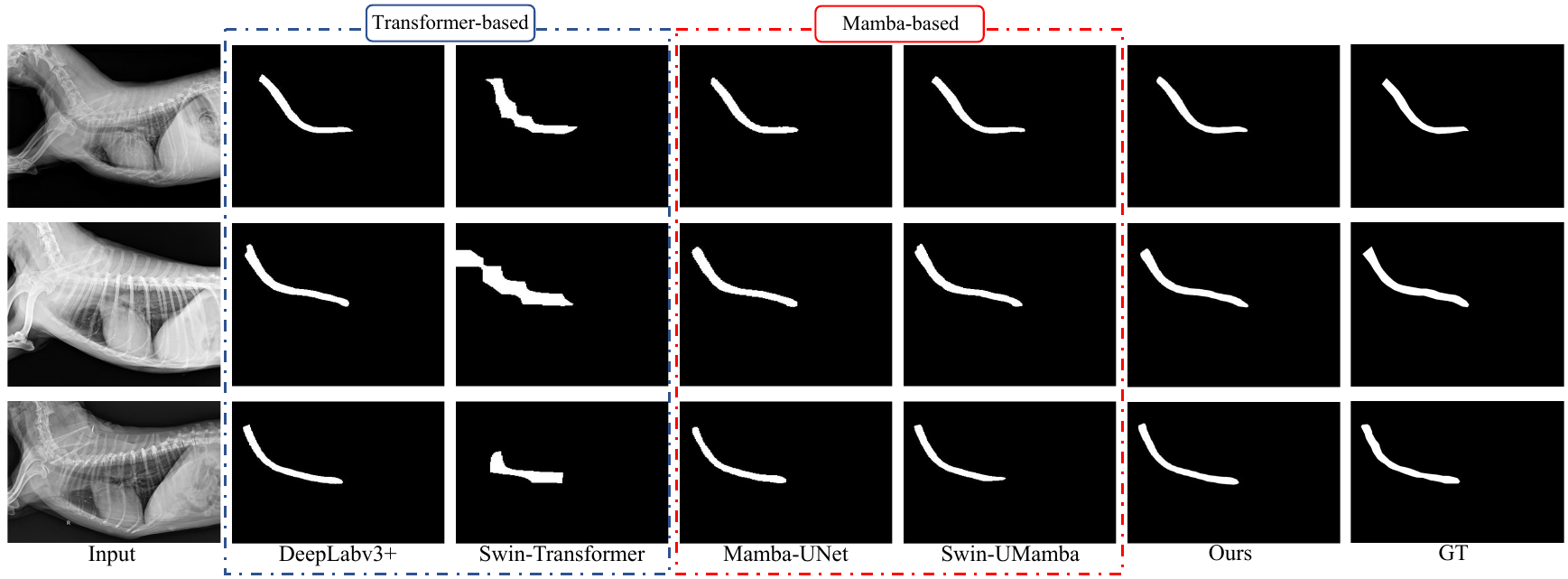}
    \caption{\textbf{Qualitative Comparison of Transformer-based and Mamba-based Segmentation Results.}}
    \label{fig:s_seg3}
\end{figure}

The overall procedure, from signal purification to the final spectral anomaly scoring, is summarized in Algorithm~\ref{alg:s_rmt_diagnosis}.

More significant performance gaps are observed against other architectural families. For instance, the general-purpose model SAM (Figure~\ref{fig:s_seg2}) and Transformer-based models like Swin-Transformer (Figure~\ref{fig:s_seg3}) largely fail on this task, producing severely fragmented or noisy results.

\begin{algorithm}[t]
\caption{Spectral Anomaly Detection and Diagnosis}
\label{alg:s_rmt_diagnosis}
\begin{algorithmic}[1]
\REQUIRE Original Image $X$, Refined Mask $\hat{M}$, Theoretical Limit $\lambda_{+}$
\ENSURE Diagnosis Label $\hat{Y}$
\STATE \textbf{Signal Purification:}
\STATE $X_{\text{focus}} \leftarrow X \odot \hat{M}$
\STATE \textbf{Statistical Modelling:}
\STATE $F_p \leftarrow \text{VLM\_Project}(X_{\text{focus}})$
\STATE $S \leftarrow \frac{1}{N} F_p^T F_p$
\STATE $\{\lambda_i\}_{i=1}^p \leftarrow \text{EigenDecomposition}(S)$
\STATE \textbf{Anomaly Scoring (SAS):}
\STATE $Score_{\text{sas}} \leftarrow 0$
\FOR{each eigenvalue $\lambda_i$}
    \IF{$\lambda_i > \lambda_{+}$}
        \STATE $Score_{\text{sas}} \leftarrow Score_{\text{sas}} + (\lambda_i - \lambda_{+})$
    \ENDIF
\ENDFOR
\STATE \textbf{Final Diagnosis:}
\STATE $\hat{Y} \leftarrow \Psi_{\text{clf}}(Score_{\text{sas}})$
\RETURN $\hat{Y}$
\end{algorithmic}
\end{algorithm}

\end{document}